\documentclass[runningheads]{llncs}

\newcounter{nb_datasets}
\usepackage{graphicx}
\usepackage{hyperref} 

\usepackage{amsmath} 
\usepackage{amssymb} 
\usepackage{mathtools} 
\usepackage{dsfont} 
\usepackage{tikz}
\graphicspath{{img/}} 
\usepackage{algorithm,algorithmic} 
\usepackage[capitalize,noabbrev,nameinlink]{cleveref}  

\newcommand{\alglinelabel}{%
  \addtocounter{ALC@line}{-1}
  \refstepcounter{ALC@line}
  \label[line]
}

\renewcommand{\qed}{\hfill$\blacksquare$} 

\begin{document}
\title{Optimizing Data Shapley Interaction Calculation from $\mathcal{O}(2^n)$ to $\mathcal{O}(t n^2)$ for KNN models 
}

\titlerunning{Shapley Interaction in Data Valuation} 

\author{Mohamed Karim Belaid\inst{1,2} \and
Dorra ElMekki\inst{3} \and
Maximilian Rabus\inst{1} \and
Eyke H\"ullermeier\inst{4}}
\authorrunning{K. Belaid et al.} 
\institute{Porsche AG, Germany \and
IDIADA GmbH, Germany \and
Passau University, Germany \and
Munich Center for Machine Learning, Germany
}


\maketitle
\begin{abstract}
With the rapid growth of data availability and usage, quantifying the added value of each training data point has become a crucial process in the field of artificial intelligence. The Shapley values have been recognized as an effective method for data valuation, enabling efficient training set summarization, acquisition, and outlier removal. In this paper, we introduce "STI-KNN", an innovative algorithm that calculates the exact pair-interaction Shapley values for KNN models in $\mathcal{O}(t n^2)$ time, which is a significant improvement over the $\mathcal{O}(2^n)$ time complexity of baseline methods. By using STI-KNN, we can efficiently and accurately evaluate the value of individual data points, leading to improved training outcomes and ultimately enhancing the effectiveness of artificial intelligence applications.


\end{abstract}

\section{Introduction}\label{sec:intro}
Data valuation, in data science, is the process that aims to quantify the added value of a train data point given a specific test dataset.
Considering that data points are sometimes expensive to acquire or difficult to label, data valuation could help in making the right decision while investing time and effort in expanding the training set, removing mislabeled points, or summarizing the training set.

\subsubsection{Shapley vs. LOO method}
Leave-one-out (LOO) is a known method for estimating the contribution of an element \cite{jia2021scalability}. Let $N$ be the training set. LOO estimates the contribution of a data point $i$ as the difference between the test score after training on $N$ and the test score after training on $N \setminus \{i\}$.
On the other hand, the Shapley method considers all subsets $S \subseteq N$. The Shapley method estimates the contribution of train point $i$ as the average difference between the test score when training on $S$ and the test score after training on $S \setminus \{i\}$. It was proven in related work that the Shapley method is more accurate in estimating the contribution of a point than LOO \cite{jia2021scalability,ghorbani2019interpretation,ghorbani2019data}. 

In previous works, researchers tackle data valuation by introducing several methods to estimate the value of train data points \cite{ghorbani2019data,jia2019efficient,kwon2021beta}. Cited methods are based on the Shapley values \cite{shapley1953quota}. Consequently, they disregard the interaction term.

\subsubsection{KNN-Shapley} \cite{jia2019efficient} computes the exact Shapley values if the model is a K-Nearest-Neighbor (KNN). Despite that KNN is a simple ML model, it is still possible to work on complex tasks like image classification thanks to the usage of pre-trained models: given a pre-trained feature extractor for images that is independent of the training set to valuate, the KNN model is trained each time on the extracted feature rather than the initial image. Moreover, Jia et al. proved that the Shapley values proposed by the KNN model are transferable to other models like Gradient Boosting.
With an execution time of $\mathcal{O}(n\log{}n)$ \cite{jia2019efficient}, KNN-Shapley is the fastest data valuation algorithm while it scales to complex tasks. Jia et al. reduced the complexity of the Shapley values by defining the valuation function as the likelihood of the right label.


Our contribution is the following:
We propose, \textbf{STI-KNN}, the first algorithm that calculates the exact pair-interaction Shapley values in $\mathcal{O}(t n^2)$ rather than $\mathcal{O}(2^n)$. STI-KNN is the first algorithm that allows studying the exact interaction on large real-world datasets. 
This research paper is the first to consider two disjoint fields: Data valuation and Interaction in Explainable AI. 
Finally, we study various cases of positive and negative data interactions using STI-KNN.

\section{Setting and Notation}

\subsection{Valuation function of KNN model}
$N$ is the training set of size $n$. \newline
$v: N \rightarrow \mathbb{R}$ is a valuation function that trains on $S \subseteq N$ and returns the test score. It varies with the model and the metric.

\cref{eq:likelihood_all_tests,eq:likelihood} are used in literature as the valuation function of KNN models \cite{jia2019efficient,jia2021scalability}. The test score is defined as the likelihood of the right label. 
\begin{equation}\label{eq:likelihood_all_tests}
    v(S) = \frac{1}{t} \sum_{y_{test} \in T}u_{y_{test}}(S)
\end{equation}
\begin{equation}\label{eq:likelihood}
    u_{y_{test}}(S) = \frac{1}{k} \sum_{i=1}^{min(k,s)} \mathds{1}_{[y_i = y_{test} ]}
\end{equation}

\noindent $T$ is the test set of size $t$.\newline
$k$ is the parameter of KNN.\newline
$y_{test} \in T$ is the label of a test data point.\newline
$y_i \in S$ is the train point label, sorted starting from the nearest to $y_{test}$.\newline
$\mathds{1}_{[y_i = y_{test} ]}$ returns 1 if both train and test labels are equal, otherwise returns 0.

\subsubsection{Example} 
\begin{figure}
\includegraphics[width=.33\textwidth]{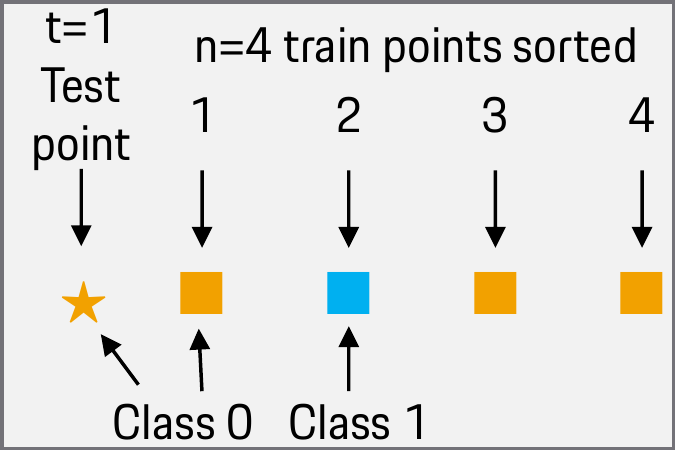}
\caption{Example of a simple dataset to explain the test score calculation for KNN model.} \label{fig:knn_example}
\end{figure}

Consider a KNN model with parameter $k=3$, $t=1$ test point denoted $y_{test}$, $n=4$ train points sorted starting from the closest to the test point, $N = \{1, 2, 3, 4\}$, See \cref{fig:knn_example}.

$v(N) = 2 / 3$ 
\begin{proof}
$v(N) = u_{y_{test}}(N)$ because the test score $v(N)$ is the average over the test scores on each $y_{test}$ but in this example, there is only one test point. \newline
$u_{y_{test}}(N) = u_{y_{test}}(\{1 ,2 ,3 \})$ because the KNN model considers only the $k=3$ closest points. Finally, there are only two data points of the same class \newline
$u_{y_{test}}(\{1,2,3 \}) = 2 / k = 2 / 3$ 
\qed \end{proof}
The calculation of the Shapley values requires training and testing the model with various training sets. In the following examples, we illustrate the calculation of the valuation function for alternative training sets:
$u_{y_{test}}(\{1\}) = 1 / 3 $ \newline 
$u_{y_{test}}(\{2\}) = 0 / 3 $ \newline 
$u_{y_{test}}(\{1 , 3 , 4\}) = 3 / 3$ 


\subsection{Shapley Taylor Interaction $\mathcal{O}(2^n)$}
$\phi_{ij}$ is the Shapley interaction value between data points $i$ and $j$~\cite{sundararajan2020shapley}.
It is the average interaction over different training sets $S \subseteq N$.  \newline
\begin{equation}\label{eq:sti_original}
    \phi_{ij}(v) = \frac{2}{n} \sum_{S \subseteq N \setminus \{i,j\}}{\frac{1}{ \binom{n-1}{s} } (v(S\cup \{i,j\}) - v(S\cup \{i\}) - v(S\cup \{j\}) + v(S))}
\end{equation}

\subsubsection{Example of a simple interaction}
\begin{figure}
\includegraphics[width=.33\textwidth]{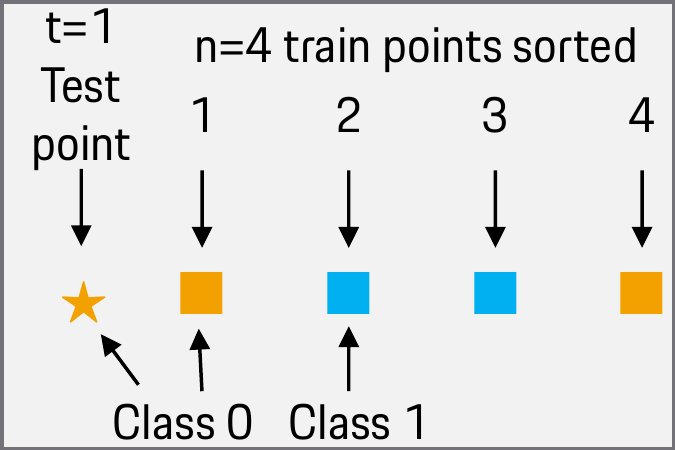}
\caption{Example of a simple dataset to explain a pair interaction between train data points.} \label{fig:interaction_example}
\end{figure}

Consider the simple dataset in \cref{fig:interaction_example}. Let $k=2$ and $N = \{1, 2, 3, 4\}$. To calculate $\phi_{1,2}$, the interaction between $i=1$ and $j=2$, we consider all subsets $S \subseteq N \setminus \{i,j\}$ and calculate the term $\phi_{i,j}$\\
$v(S\cup \{i,j\}) - v(S\cup \{i\}) - v(S\cup \{j\})$\\
For $S = \{3, 4\}$:\\ $v(S\cup \{i,j\}) - v(S\cup \{i\}) - v(S\cup \{j\}) + v(S) = 1/2 - 1/2 - 0 + 1/2 = 1/2 $ \newline
For $S = \{3\}$: $I = 1/2 - 0 - 1/2 + 0 = 0 $ \newline
For $S = \{4\}$: $I = 1/2 - 1/2 - 2/2 + 1/2 = 1/2 $ \newline
For $S = \emptyset$: $I = 1/2 - 0 - 1/2 + 0 = 0 $ \newline
Finally, $\phi_{ij}(v) = \frac{2}{n} \sum_{S \subseteq N \setminus \{i,j\}}{\frac{1}{ \binom{n-1}{s} } (v(S\cup \{i,j\}) - v(S\cup \{i\}) - v(S\cup \{j\}) + v(S))}$
$\phi_{1,2}(v) = \frac{2}{4} (\frac{1}{ \binom{4-1}{2} }\frac{1}{2} + 0 + \frac{1}{ \binom{4-1}{1} }\frac{1}{2} + 0 ) = 1 / 6$

\subsubsection{NP Complexity}
The Calculation of the matrix of interactions $\phi_{ij}$, $i \neq j$ is challenging because \cref{eq:sti_original} requires $\mathcal{O}(2^n)$ trainings of the model, each time over a subset $S \subseteq N$. Therefore, there are no real-world applications at this level~\cite{jia2021scalability}.
We propose to simplify $\phi_{ij}$ for a specific $v(S)$.

\subsubsection{Main term}
The calculation of the main terms $\phi_{ii}$ can be done in polynomial time and does not represent an issue.
For this reason, the focus of this paper is the interaction term $\phi_{ij}$.
\begin{equation} \label{eq:sti_main_term}
\phi_{ii} = v(i) - v(\emptyset)
\end{equation}

\subsubsection{Acquired linearity relative to the test set}
\cref{eq:likelihood_all_tests} involves a linearity relative to the test set: $\phi_{ij}(v)=\frac{1}{t}\sum_{y_{test}\in T}\phi_{ij}(u_{y_{test}})$. Moreover,  \cref{eq:likelihood} is linear for train sets of size $s \leq k$, which would allow simplifying the Shapley Interaction equation.

\subsubsection{Simple notation}
In the following sections, we will use $u_{y_{test}}(S)$ with only a singleton. Therefore, we simplify the notation by discarding the curved parenthesis. Note that in this specific case, the sum is discarded and the function returns only 0 or $1/k$ to express weather train point $i$ has the same label as test point $y_{test}$.
\begin{equation}\label{eq:likelihood_simple}
    u_{y_{test}}(i) = \frac{\mathds{1}_{[y_i = y_{test} ]}}{k} 
\end{equation}

\section{Method: STI-KNN $\mathcal{O}(t n^2)$}{

We study the interaction between data points by calculating the pair-interaction terms using STI \cite{sundararajan2020shapley}. The complexity of the STI algorithm (or any Shapley-based algorithm) is $\mathcal{O}(2^n)$. This hinders researchers from applying this xAI algorithm to real-world datasets. We propose STI-KNN, a faster exact calculation method adapted to the KNN model. STI-KNN is a recursive method that calculates the STI in an $\mathcal{O}(t n^2)$ execution time.
The proof is in \cref{anx:math_demo}

\subsection{Algorithm}
\begin{algorithm}[H]
\caption{STI-KNN for calculating the matrix of pair interaction Shapley values for a KNN classifier.}
\label{alg:sti_knn}
\algorithmicinput \ Training data $N = \{(x_i, y_i)\}_{i=1}^n$, test data $T = \{(x_{test,p}, y_{test,p})\}_{p=1}^{t}$, \\ Valuation function $u: N \rightarrow \mathbb{R}$ as defined in \cref{eq:likelihood_simple}, $k$ is the parameter of KNN.\\
\algorithmicoutput \ A matrix representing the pair interaction Shapley values $\{\phi_{i,j}\}_{1 \leq i,j \leq n}$
\begin{algorithmic}[1]

\FUNCTION{STI-KNN-one-test($x_{test}, y_{test}$)} \alglinelabel{alg:line:func}
\STATE $(\alpha_{1}, ..., \alpha_{n}) \gets$ Index sorting of training data starting from the closest to $x_{test}$ \alglinelabel{alg:line:sort}
    \STATE $\phi_{\alpha_{n-1},\alpha_n}(u_{y_{test}}) \gets - \frac{2(n-k)}{ n(n-1)  } u_{y_{test}}(\alpha_n)$ \alglinelabel{alg:line:n}
\FOR{$j \gets n$ to $1$} \alglinelabel{alg:line:recursive}
    \IF{$j>k+1$}
        \STATE $\phi_{\alpha_{j-2},\alpha_{j-1}}(u) \gets \phi_{\alpha_{j-1},\alpha_j}(u_{y_{test}}) + \frac{2(j - k - 1)}{(j - 2) (j - 1)}(u_{y_{test}}(\alpha_j)-u_{y_{test}}(\alpha_{j-1}))$ \alglinelabel{alg:line:j}
    \ELSE
        \STATE $\phi_{\alpha_{j-2},\alpha_{j-1}}(u) \gets \phi_{\alpha_{j-1},\alpha_j}(u_{y_{test}})$ \alglinelabel{alg:line:close}
    \ENDIF
\ENDFOR
\FOR{$j \gets 3$ to $n$} \alglinelabel{alg:line:repeat}
    \FOR{$i \gets j-2$ to $1$}
        \STATE $\phi_{\alpha_{i},\alpha_j}(u_{y_{test}}) \gets \phi_{\alpha_{i+1},\alpha_j}(u_{y_{test}})$
    \ENDFOR
\ENDFOR
\STATE \textbf{return} $\{\phi_{i,j}(u_{y_{test}})\}_{1 \leq i,j \leq  n}$ The pair-interaction matrix relative to one test point \alglinelabel{alg:line:return}
\ENDFUNCTION

\algorithmicmain
\FOR{$p \gets 1$ to $t$} \alglinelabel{alg:line:loop_t}
    \STATE $\{\phi_{i,j}(u_{y_{test,p}})\}_{1 \leq i,j \leq  n} \gets $ \textbf{STI-KNN-one-test}($x_{test,p}, y_{test,p}$)
\ENDFOR
\STATE $\{\phi_{i,j}\}_{1 \leq i,j \leq n} \gets $mean over$\ p: \{\phi_{i,j}(u_{y_{test,p}})\}_{1 \leq i,j \leq  n}$ \alglinelabel{alg:line:average}

\algorithmicendmain
\end{algorithmic}
\end{algorithm}

\subsubsection{Explaining the STI-KNN algorithm}
\label{theorem:sti_knn}
Consider the valuation function of KNN models defined in \cref{eq:likelihood_all_tests,eq:likelihood}. The Shapley-Taylor Interaction index ($\phi_{ij}$) can be calculated recursively using \cref{alg:sti_knn}.

\noindent We explain first the function \textbf{STI-KNN-one-test} that considers one test point and a train set $N = \{1,...,n\}$. This function will return $\phi_{ij}(u_{y_{test}})$ the pair interaction matrix considering only one test point.\newline
\cref{alg:line:sort}: the train points are sorted from $\alpha_{1}$ to $\alpha_{n}$, starting from the closest to the test point.\newline
\cref{alg:line:n}: calculates the pair interaction between the two points that are the most far away from $x_{test}$.
\begin{equation}\label{eq:phi_n1_n}
    \phi_{n-1,n}(u_{y_{test}}) = - \frac{2(n-k)}{ n(n-1)  } u_{y_{test}}(\alpha_n)
\end{equation}
\noindent \cref{alg:line:recursive}: The loop allows the calculation of the superdiagonal of the matrix recursively. 
\begin{equation}\label{eq:sti_phi_j2_j1}
\phi_{j-2,j-1}(u_{y_{test}}) = \begin{cases*}
      \phi_{j-1,j}(u_{y_{test}}) + \frac{2(j - k - 1)}{(j - 2) (j - 1)}(u_{y_{test}}(\alpha_j)-u_{y_{test}}(\alpha_{j-1})), & if $j>k+1$,\\
  \phi_{j-1,j}(u_{y_{test}}),                    & otherwise.
\end{cases*}
\end{equation}
\cref{alg:line:j}: The value of pair interaction changes during the recursive calculation if $u_{y_{test}}(j) \neq u_{y_{test}}(j-1)$, i.e., when the label of data point $\alpha_j$ differ from $\alpha_{j-1}$. \newline
\cref{alg:line:close}: The pair-interactions between the train points that are too close to the test point do not differ. This can be explained by the fact that the KNN model does not differentiate them. \newline
\cref{alg:line:repeat}: The elements of the superdiagonal were calculated in the steps before. In the nested loop, we calculate the remaining elements by repeating the superdiagonal. Indeed, the elements of each column of the upper triangle (without the diagonal) are equal, i.e., \newline 
$\forall a,b$ with $a<j, b<j$
\begin{equation}\label{eq:sti_columns}
\phi_{aj}(u_{y_{test}})=\phi_{bj}(u_{y_{test}})
\end{equation}
\cref{alg:line:return}: The function returns a matrix of pair interaction considering only one test point. \newline
\cref{alg:line:loop_t}: All matrices are calculated and saved. \newline
\cref{alg:line:average}: \cref{eq:phi_n1_n,eq:sti_phi_j2_j1,eq:sti_columns} calculate $\phi_{ij}(u)$ index for one test point. To obtain the pair interaction matrix $\phi_{ij}(v)$ for all test points $T$}, we average over all calculated $\phi_{ij}(u_{y_{test}})$
\begin{equation}\label{eq:sti_v}
\phi_{ij}(v)=\frac{1}{t} \sum_{y_{test} \in T}  \phi_{ij}(u_{y_{test}})
\end{equation}
\begin{proof}
    See \cref{anx:math_demo}
\qed \end{proof}

\subsection{Analysis of the Data Interaction matrix}
The STI indices are symmetric. Therefore, we study only the upper triangle, i.e., $i<j$.

\subsubsection{Unexpected Independence property} The interaction $\phi_{ij}(u_{y_{test}})$ is independent of point $i$  in the case of one test point (demonstrated in \cref{eq:sti_columns}). This is explained by the fact that KNN does not consider each point individually, but it considers only the order in which the train points come.
Nevertheless, once we average over multiple test points, $\phi_{ij}$ becomes dependent on both $i$ and $j$.

\subsubsection{The parameter $k$ has a negligible effect on the pair-interaction matrix}
Choosing to work with KNN as a surrogate model did speed up the computation. But, on the other hand, it did introduce another parameter: $k$.
We prove that, in practice, $k$ does not change the overall shape of the interaction matrix. We conduct an empirical experiment based on \arabic{nb_datasets} simple datasets described in \cref{anx:datasets}. We consider the following range for the $k$ parameters: $3 \leq k \leq 20$. For each $k_1,k_2$ in the selected range, we find that the Pearson's correlation between the two STI-KNN matrices is each time higher than $0.99$. This involves an insignificant change between pair-interaction matrices. Moreover, visual comparison of the matrices does not reveal any difference, see \cref{anx:param_k}. 

\subsubsection{Complexity} $n$ is the training set size.  $t$ is the testing set size. The complexity of the function \textbf{STI-KNN-one-test} (\cref{alg:line:func}), is $\mathcal{O}(n^2)$ because, first, it sorts the training points in $\mathcal{O}(n \log{}n)$ (\cref{alg:line:sort}), second, it calculates recursively the super-diagonal in $\mathcal{O}(n)$(\cref{alg:line:recursive}), and third it repeats the values in $\mathcal{O}(n^2)$.
The main script loop through the testing set and apply the function in $\mathcal{O}(t n^2)$ (\cref{alg:line:loop_t}), then it averages over the list of matrices by reducing the dimensions from (n,n,t) to (n,n) which costs $\mathcal{O}(t n^2)$ (\cref{alg:line:average}).
Thus, the complexity is $\mathcal{O}(t n^2)$.
\paragraph{The baseline algorithm's complexity considering $t$.}
The baseline algorithm calculates the simple Shapley values, i.e., a one-dimensional array of Shapley values representing one value per train point. It was declared with a complexity of $\mathcal{O}(n \log{}n)$ without considering the size of the testing set. While the baseline algorithm does sort the train points each time with respect to a specific test point resulting in a complexity of $\mathcal{O}(t n \log{}n)$.
\paragraph{Effect of t on the complexity.}
Depending on the use case, we can consider $t \ll n$ then the complexity of the baseline is again $\mathcal{O}(n \log{}n)$ and the complexity of STI-KNN is $\mathcal{O}(n^2)$. On the other hand, $t$ could have a significant effect on the time complexity, especially when using an 80/20 train test split as recommended in literature~\cite{gholamy201870}. Then $t \sim n$ and the baseline complexity becomes $\mathcal{O}(n^2 log n)$ while the complexity of STI-KNN becomes $\mathcal{O}(n^3)$.

\subsubsection{The STI-KNN values are approximately centered} i.e., $mean(\{\phi_{ij}\}) \approx 0$
\begin{proof}
    Consider the efficiency axiom fulfilled by STI which states that the sum of the values is equal to $a_{test}$, the test accuracy when trained on the entire train set, i.e., $\sum{\phi_{ij}} = a_{test}$. Thus,
    
    $mean(\{\phi_{ij}\}) = \frac{1}{n^2} \sum{\phi_{ij}} = \frac{a_{test}}{n^2}$
    
    Given $0 < a_{test} \leq 1$ and $n \gg 1$, $\frac{a_{test}}{n^2}  \approx 0$
\qed \end{proof}

\subsubsection{The main terms are always positive}
\begin{proof}
Consider \cref{eq:sti_main_term}: $ \phi_{ii} = v(i) - v(\emptyset) $.
$v(i)$ is defined as the likelihood \cite{jia2019efficient}, then it is always positive, and $v(\emptyset) = 0$.
\qed \end{proof}

\subsubsection{Similar pair interaction algorithms} The obtained result for STI could be applied to SII \cite{grabisch1999axiomatic}, a similar pair interaction algorithm. The only difference would be in the coefficient. For example for SII: $\phi_{n-1,n}(u_{y_{test}}) = - \frac{1}{n-1} u_{y_{test}}(\alpha_n)$. 

We further discuss the visual interpretation of the STI-KNN matrix in \cref{sec:matrix_interpretation}.

\section{Discussion: Example of Data Interaction matrices}\label{sec:matrix_interpretation}
\begin{figure}
\includegraphics[width=0.45\textwidth]{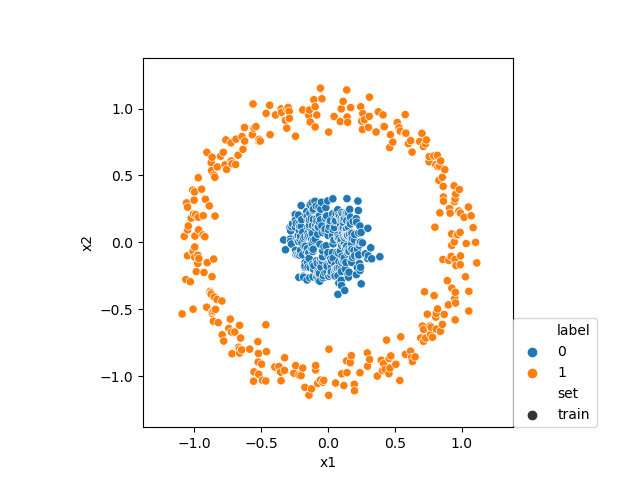}
\includegraphics[width=0.45\textwidth]{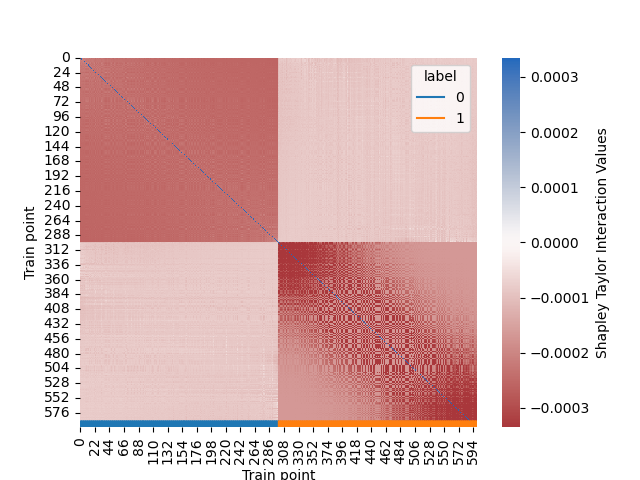}
\caption{Example of interaction with a balanced dataset.} \label{fig:circle}
\end{figure}

In this section, we analyze different types of interaction. We select the \textit{Circle dataset} \cite{pedregosa2011scikit}. Its input features are 2D points and it is a binary classification task. The distribution of the two classes represents two concentric circles, See \cref{fig:circle}. The two classes are balanced. Each class contains 300 generated points.

In the matrix, the points are first sorted by their class (0 or 1). Within each class, the points are sorted based on their input feature $x_1$, and then by $x_2$. For example, the point at index 0 is a blue point (class 0) with the smallest $x_1$ value. 

\paragraph{In-class vs. Out-of-class interaction.}
We observe that points in the same group heavily interact (negatively), while pairs of points formed by both groups almost do not interact. Clusters are clearly visible in the interaction matrix.
A pair interaction Shapley value reflects the contribution of a training point to the test score as defined initially in the valuation function. This value is the result of many factors like the correctness of the input feature (is the point an outlier?), the correctness of the target class (is the point mislabeled?), the redundancy of the training points, etc. We can emphasize one of these effects with the following variation of the circle dataset.

\paragraph{Redundancy decreases in-class interaction.}
Consider one training point $P_1$ with a positive Shapley value. With a redundant point $P_2$, both points will have the same Shapley value (symmetry axiom of the Shapley values).

\begin{figure}
\includegraphics[width=0.45\textwidth]{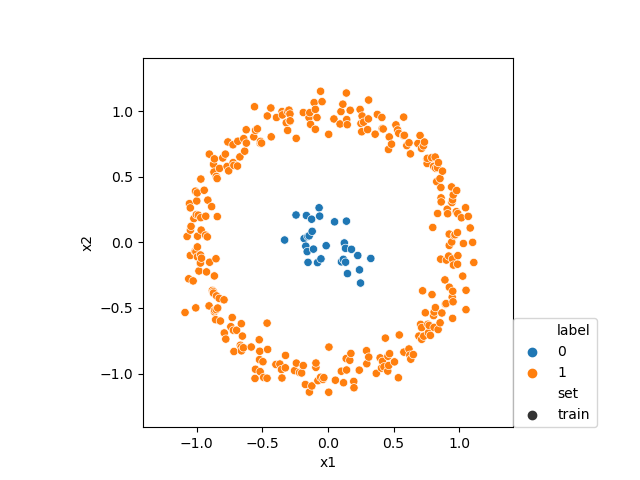}
\includegraphics[width=0.45\textwidth]{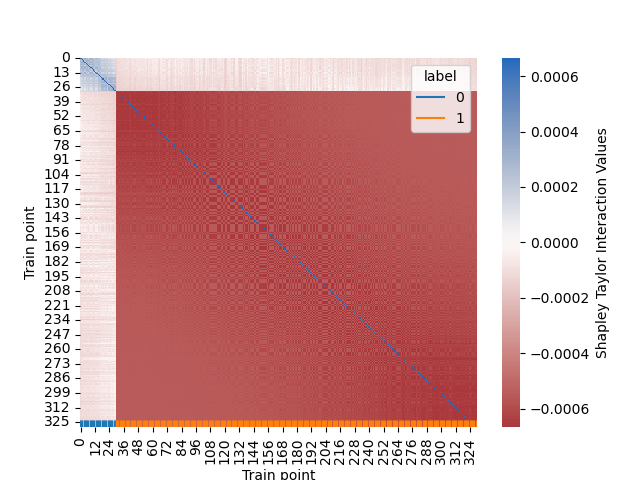}
\caption{Example of interaction with an unbalanced dataset.} \label{fig:circle_unbalance}
\end{figure}

The sum of the interaction matrix is the test accuracy (efficiency axiom of the STI). Without loss of generality, suppose the test accuracy is equal to 1. By adding redundant training points and having the same accuracy, the individual contribution of each point or each pair of points will decrease.
\cref{fig:circle_unbalance} illustrates this effect with fewer blue points and an accuracy equal to 1. The removed blue points are not perfectly overlapping (perfectly symmetric) but considering the KNN algorithm, the distance is not significant.


\begin{figure}
\includegraphics[width=0.45\textwidth]{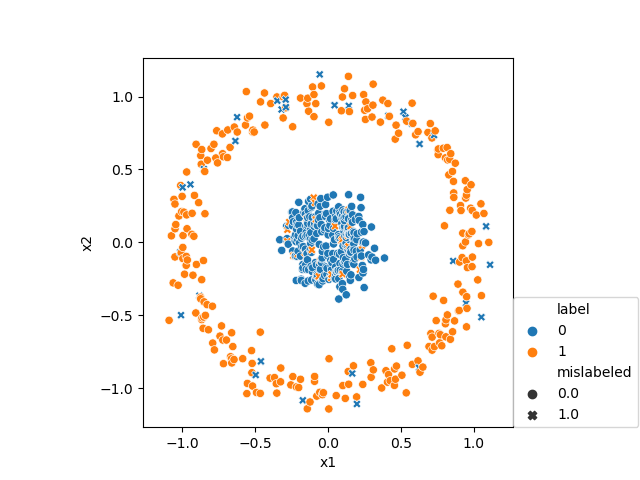} 
\includegraphics[width=0.45\textwidth]{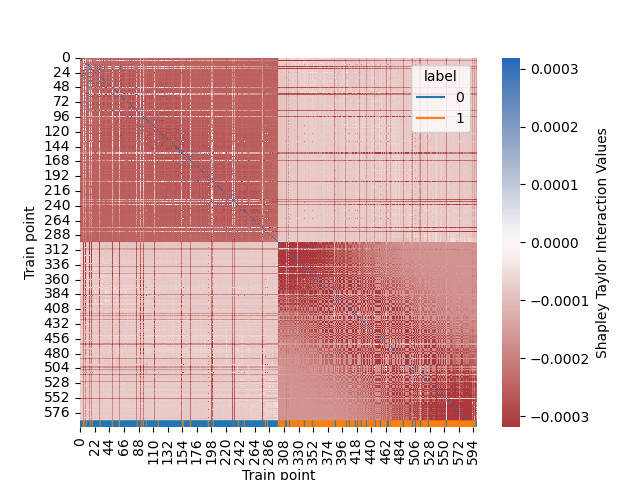} 
\caption{Example of interaction with mislabeled training points.} \label{fig:circle_mislabeled}
\end{figure}

\paragraph{Mislabeled points behave like the opposite class.}
Noisy points or mislabeled points like in \cref{fig:circle_mislabeled} tend to behave differently compared to the majority of the points in the same class. The interaction matrix in \cref{fig:circle_mislabeled} (right) helps to identify mislabeled points as their pattern corresponds more to the opposite class.






\section{Conclusions and Future Work}
\label{sec:conlusion}
The considerable value of data creation in industries such as medicine and automotive has led to an increased focus on data generation, crowdsourcing, and simulation techniques.
These latter provide a variety of benefits such as improving decision-making and enhancing efficiency. On the other hand, these techniques raise the urgent need for data valuation to confirm the value of data and its role in business and society. The value of data can be difficult to quantify and is often influenced by factors such as quality, relevance, and accessibility. Thus, it is crucial to develop methods for accurately valuating data and maximizing its contribution to AI models.

This paper bridges the gap between Shapley interaction and data valuation by introducing STI-KNN, the first algorithm that calculates the exact pair-interaction Shapley values with a complexity of $\mathcal{O}(t n^2)$ rather than $\mathcal{O}(2^n)$.

\bibliographystyle{splncs04} 
\bibliography{refs} 

\newpage
\appendix
\section{Lemmas and Proof of \cref{theorem:sti_knn}}\label{anx:math_demo}
\begin{figure}[!h]
\begin{tikzpicture}
\matrix [column sep=3mm, row sep=9mm] {
\node {\cref{anx:sub:proof_phi_n1_n}:}; & &   \node (a1)  {\cref{eq:sti3_restricted_sum}}; &   \node (a2)  {\cref{eq:phi_n1_n}}; \\
\node {\cref{anx:sub:proof_reccursive}:}; &  \node (b1) {\cref{eq:sti_diff}}; &  \node (b2) {\cref{eq:sti_diff_restricted_sum}};  &  \node (b3) {\cref{eq:sti_phi_j2_j1}}; &  \node (b4) {\cref{alg:sti_knn}}; \\  
\node {\cref{anx:sub:proof_columns}:}; &                                    &   &  \node (c1) {\cref{eq:sti_columns}}; \\
};
\draw[->, thick] (a1) -- (a2);
\draw[->, thick] (b1) -- (b2);
\draw[->, thick] (b2) -- (b3);

\draw[->, thick] (b1) -- (c1);

\draw[->, thick] (a2) -- (b4);
\draw[->, thick] (b3) -- (b4);
\draw[->, thick] (c1) -- (b4);
\end{tikzpicture}
\caption{The steps for the proof of \cref{alg:sti_knn}.}\label{fig:proof_flow}
\end{figure}
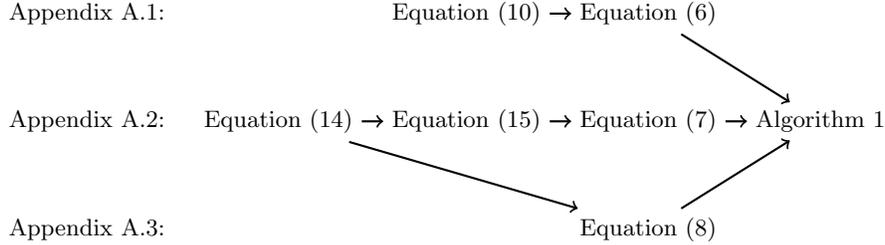

\cref{alg:sti_knn} is composed of three main equations (\cref{eq:phi_n1_n,eq:sti_phi_j2_j1,eq:sti_columns}).
The proof is divided into three subsections demonstrating each of these three main equations.
\cref{fig:proof_flow} summarizes the flow of the proofs involving all lemmas and equations.

\subsubsection{Simplify notations}
In the following, we simplify unions between sets by writing for example $v(Sij)$ rather than $v(S\cup \{i,j\})$.
Without loss of generality, we suppose that train data points are sorted starting from the closest to $y_{test}$. Sorted train data points were denoted $N = \{\alpha_1 , ... , \alpha_n \}$. But we simplify the notation to $N = \{1 , ... , n \}$.
Finally, since in all demonstrations, we will work with the function $u_{y_{test}}$ from \cref{eq:likelihood_simple}, we simplify its notation to $u$.

\subsection{Proof of the last term}\label{anx:sub:proof_phi_n1_n}

The proof of \cref{alg:sti_knn} starts with the Proof of \cref{eq:sti3_restricted_sum}. 

\begin{lemma}[STI Equation with restricted double sum] 
$ \forall i \neq j $

\begin{equation}\label{eq:sti3_restricted_sum}
\phi_{ij}(u) =  \frac{2}{n} \sum_{s=k-1}^{n-2}{  \frac{1}{ \binom{n-1}{s} } \sum_{S, |S|=s,\\ S \subseteq N \setminus \{i,j\}}(u(Sij) - u(Si) - u(Sj) + u(S))}
\end{equation}
\end{lemma}
\begin{proof}[Proof of \cref{eq:sti3_restricted_sum}]\label{proof:sti3_restricted_sum}
The sum in \cref{eq:sti_original} can be divided into two sums: one over the size of the sets $S$ and one over the content of sets $S$:
\begin{equation}\label{eq:sti_double_sum}
\phi_{ij}(u)  =  \frac{2}{n} \sum_{s=0}^{n-2}{  \frac{1}{ \binom{n-1}{s} } \sum_{S, |S|=s, S \subseteq N \setminus \{i,j\}}(v(Sij) - v(Si) - v(Sj) + v(S))}
\end{equation}
For all $s\leq k-2$, the total size of each of $Sij$, $Si$, $Sj$, and $S$ is below $k$. Therefore, $u$ becomes linear:\newline
$u(Sij) - u(Si) - u(Sj) + u(S)\\
= u(S) + u(i) + u(j) - u(S) - u(i) - u(S) - u(j) + u(S)\\
= 0$\newline
Therefore the first sum can be limited to $k-1\leq s\leq n-2$
\qed \end{proof}

We remind the \cref{eq:phi_n1_n} with the simplified notation:
\begin{equation}
    \phi_{n-1,n}(u) = - \frac{2(n-k)}{ n(n-1)  } u(n)
    \tag{\ref{eq:phi_n1_n}}
\end{equation}

\begin{proof}[Proof of \cref{eq:phi_n1_n}]
Consider \cref{eq:sti3_restricted_sum} with $i=n-1$ and $j=n$
\begin{equation}
\phi_{n-1,n}(u) =  \frac{2}{n} \sum_{s=k-1}^{n-2}{  \frac{1}{ \binom{n-1}{s} } \sum_{S, |S|=s, S \subseteq N \setminus \{n-1,n\}}(u(S \cup \{n-1, n\}) - u(S \cup \{n-1\}) - u(S \cup \{n\}) + u(S))}
\end{equation}
We distinguish two cases depending on the size of S.
\begin{enumerate}
    \item $s>=k$\\
    $u(S \cup \{n-1, n\}) = u(S \cup \{n-1\}) = u(S \cup \{n\}) = u(S)$ Thus \\
    $u(S \cup \{n-1, n\}) - u(S \cup \{n-1\}) - u(S \cup \{n\}) + u(S) = 0$

    \item $s=k-1$\\
    $u(S \cup \{n-1, n\}) = u(S \cup \{n-1\})$
    and 
    $u(S \cup \{n\}) = u(S) + u(n)$
    Thus
    \begin{equation}
    \begin{split}
    \phi_{n-1,n}(u) & = \frac{2}{n} \frac{1}{ \binom{n-1}{k-1}} \sum_{S, |S|=k-1, S \subseteq N \setminus n-1,n}(-u(n))\\
    & =\frac{2}{n} \frac{1}{ \binom{n-1}{k-1}} \binom{n-2}{k-1} (-u(n))\\
    & = - \frac{2(n-k)}{ n(n-1)  } u(n)
    \end{split}
    \end{equation}
\end{enumerate}

\qed \end{proof}

\subsection{Proof of the recursive interaction index}\label{anx:sub:proof_reccursive}

\begin{lemma}[Difference between interaction indices] 
$ \forall i \neq j \neq q$ (without any order)
\begin{equation}\label{eq:sti_diff}
    \phi_{ij}(u) - \phi_{iq}(u)
    =\frac{2}{n} \sum_{S \subseteq N \setminus \{i,j,q\}}{(\frac{1}{ \binom{n-1}{s} } + \frac{1}{ \binom{n-1}{s+1} } ) (u(Sij) - u(Siq) - u(Sj) + u(Sq))}
\end{equation}
\end{lemma}

\begin{proof}[Proof of \cref{eq:sti_diff}]
We use \cref{eq:sti_original}:

$\phi_{ij}(u) - \phi_{iq}(u) \\
= \frac{2}{n} \sum_{S \subseteq N \setminus \{i,j\}}{\frac{1}{ \binom{n-1}{s} } (u(Sij) - u(Si) - u(Sj) + u(S))} - \\
\frac{2}{n} \sum_{S \subseteq N \setminus \{i,q\}}{\frac{1}{ \binom{n-1}{s} } (u(Siq) - u(Si) - u(Sq) + u(S))} \\ \\
=\frac{2}{n} \sum_{S \subseteq N \setminus \{i,j,q\}}{\frac{1}{ \binom{n-1}{s} } (u(Sij) - u(Siq) - u(Sj) + u(Sq))} + \\
\frac{2}{n} \sum_{S \subseteq N \setminus \{i,j,q\}}{\frac{1}{ \binom{n-1}{s+1} } (u(Sijq) - u(Siq) - u(Sjq) + u(Sq))} - \\
(u(Sijq) - u(Sij) - u(Sjq) + u(Sj)) \\ \\
= \frac{2}{n} \sum_{S \subseteq N \setminus \{i,j,q\}}{\frac{1}{ \binom{n-1}{s} } (u(Sij) - u(Siq) - u(Sj) + u(Sq))} + \\
\frac{2}{n} \sum_{S \subseteq N \setminus \{i,j,q\}}{\frac{1}{ \binom{n-1}{s+1} } (u(Sij) - u(Siq) - u(Sj) + u(Sq))}$

Finally, we factor the coefficients:

$\phi_{ij}(u) - \phi_{iq}(u) \\
=\frac{2}{n} \sum_{S \subseteq N \setminus \{i,j,q\}}{(\frac{1}{ \binom{n-1}{s} } + \frac{1}{ \binom{n-1}{s+1} } )(u(Sij) - u(Siq) - u(Sj) + u(Sq))} $
\qed \end{proof}

\begin{lemma}[Difference between interaction indices with a restricted sum] 
$ \forall i \neq j \neq q$ (without any order)
\begin{equation}\label{eq:sti_diff_restricted_sum}
    \phi_{ij}(u) - \phi_{iq}(u)
    =\frac{2}{n} \sum_{s=k-1}^{n-3} {(\frac{1}{ \binom{n-1}{s} } + \frac{1}{ \binom{n-1}{s+1} } ) \sum_{|S|=s,S \subseteq N \setminus \{i,j,q\}} (u(Sij) - u(Siq) - u(Sj) + u(Sq))} 
\end{equation}
\end{lemma}
\begin{proof}[proof of \cref{eq:sti_diff_restricted_sum}]
    using the same steps as in the \cref{proof:sti3_restricted_sum}: consider the case where $s<k-1$. The valuation function becomes linear. The sum is zero.

\qed \end{proof}

We remind \cref{eq:sti_phi_j2_j1} with a simplified notation:
\begin{equation}
\phi_{j-2,j-1}(u) = \begin{cases*}
      \phi_{j-1,j}(u) + \frac{2(j - k - 1)}{(j - 2) (j - 1)}(u(j)-u(j-1)), & if $j>k+1$,\\
  \phi_{j-1,j}(u),                    & otherwise.
\end{cases*}
\tag{\ref{eq:sti_phi_j2_j1}}
\end{equation}

\begin{proof}[Proof of \cref{eq:sti_phi_j2_j1}]

Consider \cref{eq:sti_diff_restricted_sum}. The STI indices are symmetric\cite{sundararajan2020shapley}. \newline
Thus $\phi_{iq}(u)  = \phi_{qi}(u)$.

\noindent We replace the train point $i$ by $j-1$, and replace $q$ by $j-2$.

\noindent $\phi_{j-1, j}(u) - \phi_{j-1, j-2}(u) \\
=\frac{2}{n} \sum_{s=k-1}^{n-3} (\frac{1}{ \binom{n-1}{s} } + \frac{1}{ \binom{n-1}{s+1} } ) \sum_{|S|=s,S \subseteq N \setminus \{ j-1, j, j-2 \}} ( \\
u(S \cup \{j-1, j\}) - u(S \cup \{j-2, j-1\}) - u(S \cup \{j\}) + u(S \cup \{j-2\})) $

We split $S$ into two sets $S_1$ and $S_2$ ($S_1 \cup S_2 = S$) while $S_1$ contains only points closer to the test points than $j-2$ i.e. $S_1 \subseteq \{1,2,...,j-3\}$ and $S_2 \subseteq \{j+1, ..., n\}$. $S_1$ is of size $s_1$.
$0=<s_1\leq j-3$. 
We distinguish three cases depending on the size $s_1$:
\begin{enumerate}
    \item $s_1>=k$ \\
    $u(S \cup \{j-1, j\}) - u(S \cup \{j-2, j-1\}) - u(S \cup \{j\}) + u(S \cup \{j-2\}) = 0$ because $u$ becomes linear.
    \item $s_1=k-1$ \\
    $u(S \cup \{j-1, j\}) - u(S \cup \{j-2, j-1\}) - u(S \cup \{j\}) + u(S \cup \{j-2\}) \\
    =u(S_1 \cup \{j-1\}) - u(S_1 \cup \{j-2\}) - u(S_1 \cup \{j\}) + u(S_1 \cup \{j-2\})$ all sets becomes of size $k$. Thus $u$ becomes linear. \\
    $=u(j-1) - u(j) $
    This case exist only if $s_1 \leq j-3$ i.e. $k-1 \leq j-3$
    \item $s_1\leq k-2$ \\
    $u(S \cup \{j-1, j\}) - u(S \cup \{j-2, j-1\}) = u(j) - u(j-2)$ \\

    $- u(S \cup \{j\}) + u(S \cup \{j-2\}) = - u(j) + u(j-2)$ \\
    Thus \\
    $u(S \cup \{j-1, j\}) - u(S \cup \{j-2, j-1\}) - u(S \cup \{j\}) + u(S \cup \{j-2\})= 0$

\end{enumerate}
\noindent Combining the three cases, the second sum can be restrained to $s_1=k-1$ because it is the only case that doesn't sum to 0. \\

$\phi_{j-1, j}(u) - \phi_{j-1, j-2}(u) \\
=\frac{2}{n} \sum_{s=k-1}^{n-3} {(\frac{1}{ \binom{n-1}{s} } + \frac{1}{ \binom{n-1}{s+1} } ) \sum_{S,s_1=k-1,|S|=s,S \subseteq N \setminus \{ j-1, j, j-2 \}} (u(j-1) - u(j) )} $ \\

The sum over $S$ becomes independent of $S$. Let $C$ be its size. \\
$C = |\{S,s_1=k-1,|S|=s,S \subseteq N \setminus j-2, j-1, j\}_{s=k-1}^{n-3}|$

We distinguish two cases:
\begin{enumerate}
    \item $j-2\leq k-1$ \\
    $S_1$ can not contain $k-1$ points. Thus, $C=0$
    \item $j-2>k-1$
   
    $s_2 = s-s_1 = s-k+1$
    
    $S_1$ takes $k-1$ points from $j-3$ points
    
    $S_2$ takes $s-k+1$  points from $n-3-(j-3) = n-j$

\end{enumerate}
We combine the two cases. Thus:
$C = \mathds{1}[j-2>k-1]\binom{j-3}{k-1}\binom{n-j}{s-k+1}$

\noindent We replace the second sum in the main equation:

$\phi_{j-1, j}(u) - \phi_{j-1, j-2}(u) \\
=\frac{2}{n} \sum_{s=k-1}^{n-3} {(\frac{1}{ \binom{n-1}{s} } + \frac{1}{ \binom{n-1}{s+1} } ) \mathds{1}[j>k+1]\binom{j-3}{k-1}\binom{n-j}{s-k+1} (u(j-1) - u(j) )} $ 


\noindent Let's simplify the term $D$ defined as \\
$D := \sum_{s=k-1}^{n-3}
(\frac{\binom{j-3}{k-1} \binom{n-j}{s-k+1}}{ \binom{n-1}{s} } + 
\frac{\binom{j-3}{k-1} \binom{n-j}{s-k+1}}{ \binom{n-1}{s+1} } )  $
\begin{enumerate}
    \item First let's simplify $D1 := \sum_{s=k-1}^{n-3}
(\frac{\binom{j-3}{k-1} \binom{n-j}{s-k+1}}{ \binom{n-1}{s} }$

We expand the first binomial using the following binomial identity \cite{jia2019efficient}:

$\binom{j-2-1}{k-1} =\binom{j-1-1}{k-1} \frac{j-2-(k-1)}{j-2}=\binom{j-1}{k-1} \frac{j-1-(k-1)}{j-1}\frac{j-2-(k-1)}{j-2} $

Thus 

$D1 = \frac{j-1-(k-1)}{j-1}\frac{j-2-(k-1)}{j-2} \sum_{s=k-1}^{n-3}
\frac{\binom{j-1}{k-1} \binom{n-j}{s-k+1}}{ \binom{n-1}{s} }  $

We change the variable by introducing $a,b,u,v$: 

$a := s-k+1$

Since $k-1\leq s\leq n-3$ Then $0\leq a\leq n-k-2$

$b := k-1$ Thus $a+b=s$

$u := j-1$

$v := n-j$ Thus $u+v=j-1 + n-j = n-1$

The upper bound of the sum is replaced as follows: $n-k-2 = v+j-k-2$

$D1 = \frac{j-1-(k-1)}{j-1}\frac{j-2-(k-1)}{j-2} \sum_{a=0}^{v+j-k-2}
\frac{\binom{u}{b} \binom{v}{a}}{ \binom{u+v}{a+b} } $

The upper bound of the sum is pushed to v by adding binomial terms equal to zero.

$D1 = \frac{j-1-(k-1)}{j-1}\frac{j-2-(k-1)}{j-2} \sum_{a=0}^{v}
\frac{\binom{u}{b} \binom{v}{a}}{ \binom{u+v}{a+b} }$

The following binomial identity is used: $\sum_{a=0}^{v}
\frac{\binom{u}{b} \binom{v}{a}}{ \binom{u+v}{a+b} } = \frac{u+v+1}{u+1}$

$D1 =\frac{j-1-(k-1)}{j-1}\frac{j-2-(k-1)}{j-2} \frac{u+v+1}{u+1}\\
=\frac{j-1-(k-1)}{j-1}\frac{j-2-(k-1)}{j-2} \frac{n}{j}$ 

\item A similar proof is applied for the second term

$D2 := \sum_{s=k-1}^{n-3}
\frac{\binom{j-3}{k-1} \binom{n-j}{s-k+1}}{ \binom{n-1}{s+1} } \\
= \frac{k}{j-1}\frac{j-2-(k-1)}{j-2} \sum_{s=k-1}^{n-3}
\frac{\binom{j-1}{k} \binom{n-j}{s-k+1}}{ \binom{n-1}{s+1} } $








$= \frac{k}{j-1}\frac{j-2-(k-1)}{j-2} \frac{n}{ j }$

\end{enumerate}

\noindent Finally

\noindent $D = D1 + D2 = \frac{n(j - k - 1)}{(j - 2) (j - 1)}$
\qed \end{proof}

\subsection{Proof of the equality of the columns}\label{anx:sub:proof_columns}

\begin{proof}[Proof of \cref{eq:sti_columns}]

\noindent Consider \cref{eq:sti_diff} \\
First, we flip the indices using symmetry: 
$\phi_{ij}(u) = \phi_{ji}(u) $ and $ \phi_{iq}(u)  = \phi_{qi}(u) $

\noindent Then, we replace the variables as follows: \\
$j$ becomes $i-1$. $i$ becomes $j$. $q$ becomes $i$. \\
Thus

$\phi_{i-1,j}(u) - \phi_{ij}(u) \\
=\frac{2}{n} \sum_{S \subseteq N \setminus \{i-1,i,j\}}{(\frac{1}{ \binom{n-1}{s} } + \frac{1}{ \binom{n-1}{s+1} } )(u(S \cup \{i-1,j\}) - u(S \cup \{i,j\}) - u(S \cup \{i-1\}) + u(S \cup \{i\}))} $ with $i<j$

We split $S$ into $S_1, S_2, S_3$ such that $S_1 \cup S_2 \cup S_3 = S$ and 

$S_1 \subseteq \{1,2,...,i-2\}$

$S_2 \subseteq \{i+1, ..., j-1\}$

$S_3 \subseteq \{j+1, ..., n\}$

$\phi_{i-1,j}(u) - \phi_{ij}(u) \\
=\frac{2}{n} \sum_{S \subseteq N \setminus \{i-1,i,j\}}(\frac{1}{ \binom{n-1}{s} } + \frac{1}{ \binom{n-1}{s+1} } )( 
u(S_1 \cup \{i-1\} \cup S_2 \cup \{j\} \cup S_3) \\
- u(S_1 \cup \{i\} \cup S_2 \cup \{j\} \cup S_3) 
- u(S_1  \cup \{i-1\} \cup S_2 \cup S_3) 
+ u(S_1 \cup \{i\} \cup S_2 \cup S_3)
) $ \\
We distinguish two cases:
\begin{enumerate}
    \item $s_1>=k$ \\
    all valuation function are equal to $u(S_1)$. Then the sum is zero.
    \item $s_1 \leq k-1$ \\
    
    
    $ - u(S \cup \{i-1\}) + u(S \cup \{i\}) = u(i) - u(i-1) $
    
    
    $ u(S \cup \{i-1,j\}) - u(S \cup \{i,j\}) = u(i-1) - u(i)$

\end{enumerate}

\noindent Thus, the sum is zero for all cases.

for a fixed $j$, 
$ \forall i<j$ 
$\phi_{i-1,j} - \phi_{ij}(u) = 0$

Recursively, considering a fixed $j$, $\forall i<j$ $\phi_{ij}$ are all equal.
\qed \end{proof}

\begin{corollary}
    The standard deviation of the STI-KNN values is inversely proportional to $k$.
\end{corollary}
\begin{proof}
    The term $\phi_{n-1,n}$ is affected by k. But since $\phi_{n-1,n}$ is propagated to all $\phi_{ij}$, the mean value will change accordingly. Thus, this term does not affect the standard deviation.
    
    $\phi_{j-2, j}$ is affected by $k$ only if $u(j) \neq u(j-1)$, i.e., when the labels of point $j$ and $j-1$ are different. In that context, $u(j) - u(j-1) = \pm \frac{1}{k}$. Thus

    $\phi_{j-2, j-1} =\phi_{j-1, j} + \frac{2}{(j - 2)k} - \frac{2}{(j - 2)(j - 1)}$

    The series $\phi_{j-2, j-1}$ is varying because of two terms; one of them is inversely proportional to $k$.
    
\qed \end{proof}

\clearpage

\section{Varying parameter $k$}\label{anx:param_k}
\begin{figure}[h] \begin{center}
\includegraphics[width=.45\columnwidth]{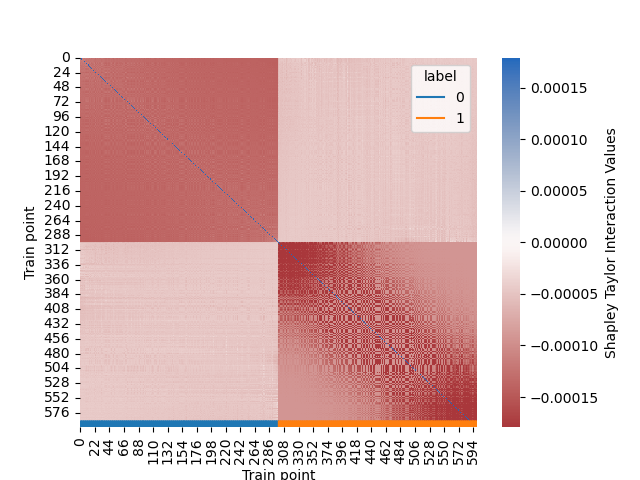}
\includegraphics[width=.45\columnwidth]{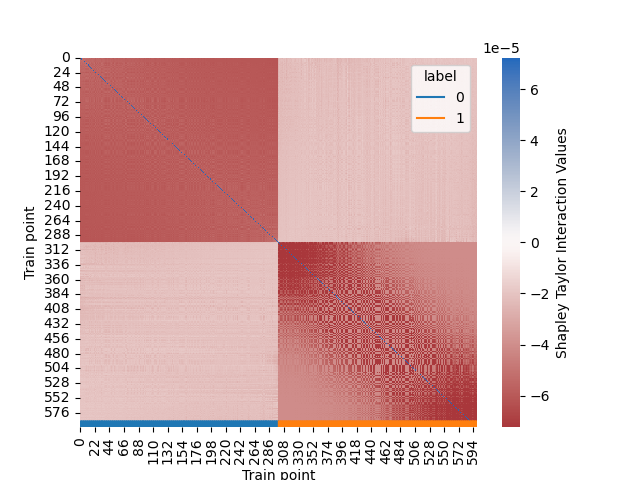}
\caption{STI-KNN pair-interaction. Circle Dataset $k = 9$ and $k = 20$}
\label{fig:anx_param_k_circle}
\end{center} \vskip -0.2in \end{figure}

\begin{figure}[!h]   \begin{center}
\includegraphics[width=.45\columnwidth]{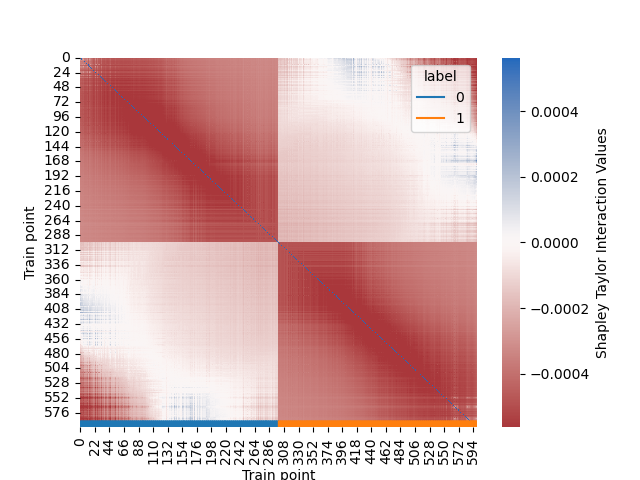}
\includegraphics[width=.45\columnwidth]{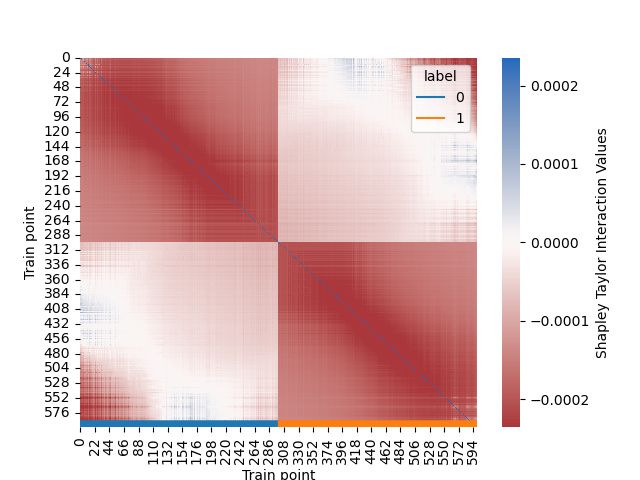}
\caption{STI-KNN pair-interaction. Moon Dataset. $k = 3$ and $k = 7$}
\label{fig:anx_param_k_moon}
\end{center} \vskip -0.2in \end{figure}

\begin{figure}[!h]   \begin{center}
\includegraphics[width=.45\columnwidth]{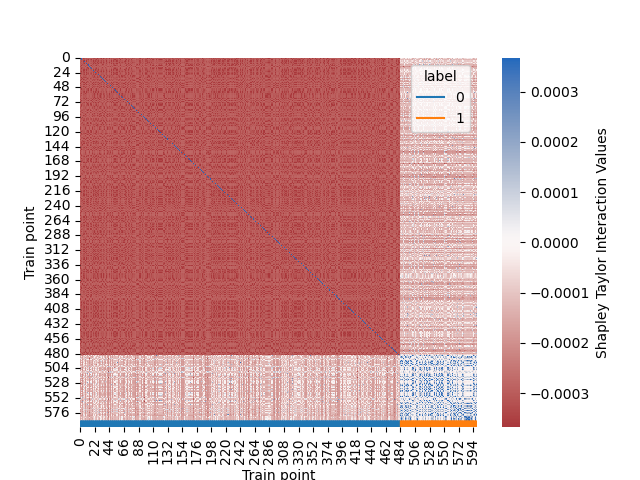}
\includegraphics[width=.45\columnwidth]{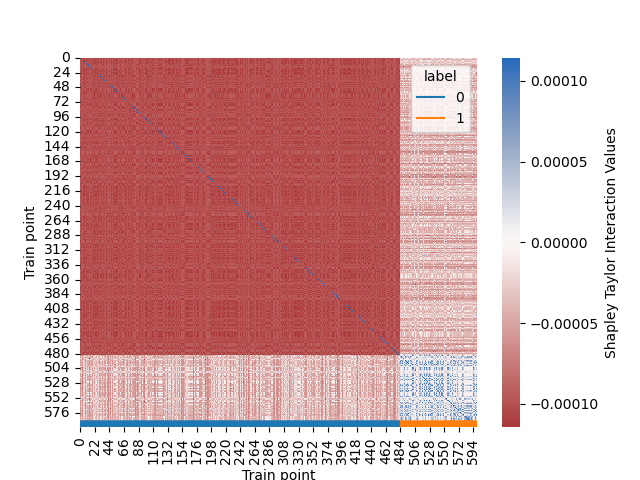}
\caption{STI-KNN pair-interaction. Click Dataset $k = 5$ and $k = 15$}
\label{fig:anx_param_k_click}
\end{center} \vskip -0.2in \end{figure}

\begin{figure}[!h]   \begin{center}
\includegraphics[width=.45\columnwidth]{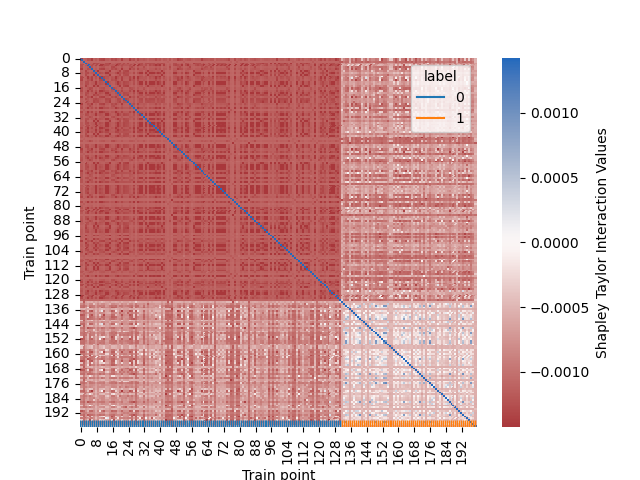}
\includegraphics[width=.45\columnwidth]{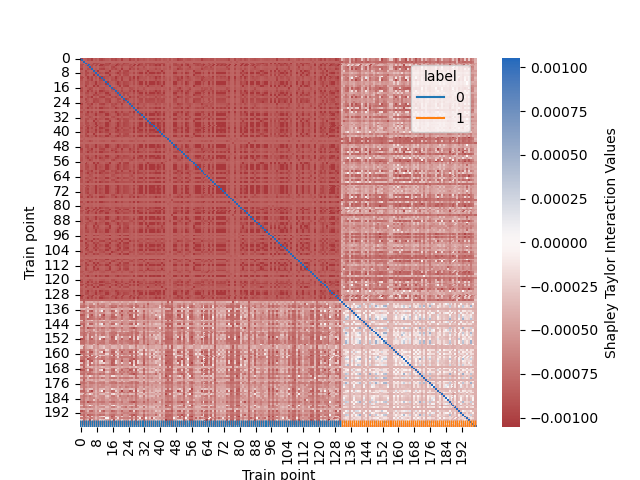}
\caption{STI-KNN pair-interaction. MonksV2 Dataset $k = 3$ and $k = 4$}
\label{fig:anx_param_k_monksv2}
\end{center} \vskip -0.2in \end{figure}

The parameter $k$ might introduce an additional workload in fine-tuning the surrogate model and in interpreting the pair-interaction values accordingly. In this section, we study the influence of the parameter $k$ visually.
\cref{fig:anx_param_k_circle} to \ref{fig:anx_param_k_monksv2} are some examples of pair interaction matrices. For each example dataset, we sample $k_1$ and $k2$ parameters to visually compare the matrices. We notice that the correlation between matrices (flattened) is above 0.99 while varying $k$. Thus, the relative change between the values is negligible. The only significant difference is the scale of the STI values.

The train points are sorted by class, then by the first input feature, then by the second input feature. For example, the Circle dataset is sorted by class then $x_1$ then $x_2$.

\section{Dataset sources} \label{anx:datasets}

\begin{table*}[t]
\caption{Evaluation datasets.}
\label{tab:datasets}
\vskip 0.15in \begin{center} \begin{small} \begin{sc}
\begin{tabular}{lll}
\hline
{} &                    Used in related work &                      Source \\
Name         &                                         &                             \\
\hline
APSFailure   &                     \cite{kwon2021beta} &          openml.org/d/41138 \\
CPU          &                     \cite{kwon2021beta} &            openml.org/d/761 \\
Circle       &                                         &  \cite{pedregosa2011scikit} \\
Click        &                     \cite{kwon2021beta} &           openml.org/d/1218 \\
CreditCard   &                     \cite{kwon2021beta} &             openml.org/d/31 \\
FashionMnist &  \cite{yoon2020data,jia2021scalability,kwon2021beta} &      \cite{xiao2017fashion} \\
Flower       &                                  \cite{yoon2020data} &          openml.org/d/43839 \\
MonksV2      &                                         &            openml.org/d/334 \\
Moon         &                                         &  \cite{pedregosa2011scikit} \\
Phoneme      &                     \cite{kwon2021beta} &           openml.org/d/1489 \\
Planes2D     &                     \cite{kwon2021beta} &            openml.org/d/727 \\
Pol          &                     \cite{kwon2021beta} &            openml.org/d/722 \\
SteelPlates  &                                         &          openml.org/d/40982 \\
TicTacToe    &                                         &             openml.org/d/50 \\
Transfusion  &                                         &           openml.org/d/1489 \\
Wind         &                     \cite{kwon2021beta} &            openml.org/d/847 \\
\hline
\end{tabular} \end{sc} \end{small} \end{center} \vskip -0.1in \end{table*}





\end{document}